\documentclass{article}

\ifdefined\XeTeXversion
  \usepackage[T1]{fontenc}
\fi
\usepackage{iclr2027_conference,times}

\usepackage{amsmath,amsfonts,bm}

\def\eqref#1{equation~\ref{#1}}

\def\1{\bm{1}}

\DeclareMathAlphabet{\mathsfit}{\encodingdefault}{\sfdefault}{m}{sl}
\SetMathAlphabet{\mathsfit}{bold}{\encodingdefault}{\sfdefault}{bx}{n}

\usepackage{amssymb}
\usepackage{booktabs}
\usepackage{tabularx}
\usepackage{array}
\usepackage{graphicx}
\usepackage{capt-of}
\usepackage{placeins}
\usepackage{xcolor}
\usepackage{colortbl}
\definecolor{rwmshade}{gray}{0.93}
\usepackage{microtype}
\usepackage{hyperref}
\usepackage{url}

\hypersetup{
    colorlinks=true,
    linkcolor=blue!50!black,
    citecolor=blue!50!black,
    urlcolor=blue!50!black,
    pdftitle={Language Models as AI Research World Models},
    pdfauthor={Zijun Wang, Zewen Liu, Minhua Lin, Zhaotian Weng, Zhan Shi, Bing He, Yisi Sang, Dakuo Wang, Benoit Dumoulin, Wei Jin, Yuyin Zhou, Cihang Xie, Hanqing Lu}
}

\title{Language Models as AI Research World Models}
\author{%
{Zijun Wang$^{1,2,*}$, Zewen Liu$^{1,3,*}$, Minhua Lin$^{1,4}$, Zhaotian Weng$^{1,5}$, Zhan Shi$^{1}$\\
Bing He$^{1}$, Yisi Sang$^{1}$, Dakuo Wang$^{1}$, Benoit Dumoulin$^{1}$}, Wei Jin$^{3}$, Yuyin Zhou$^{2}$\\
Cihang Xie$^{2}$, Hanqing Lu$^{1}$\\
{$^{1}$Amazon\ \ $^{2}$UC Santa Cruz\ \  $^{3}$Emory University\ \  $^{4}$Pennsylvania State University\ \  $^{5}$UC Santa Barbara\\
{\footnotesize $^{*}$Equal contribution.}}\\
\amazontitleemail{luhanqin@amazon.com}%
}

\iclrfinalcopy

\usepackage[boxed]{amazontitle}
\amazontitlesetup{
  author-layout={2,1,1},
  affiliation={},
  title-break-after={:},
  logo={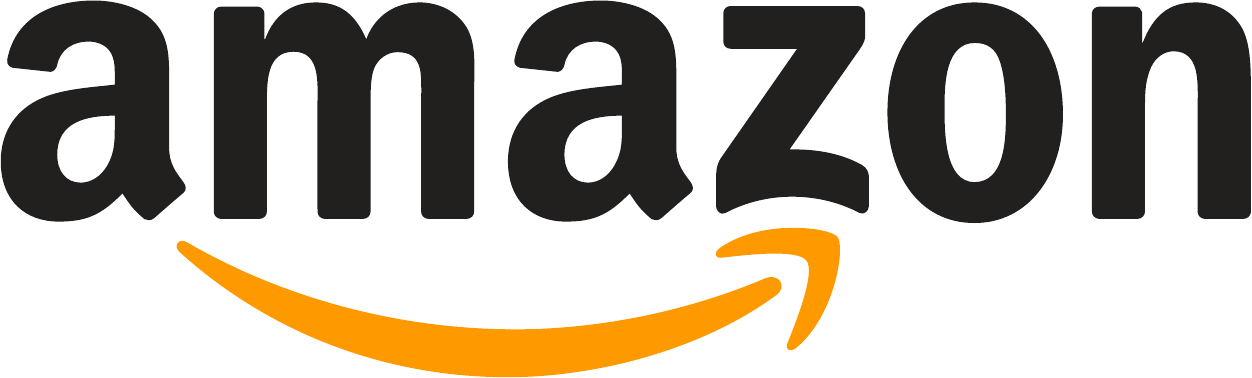},
  correspondence={\amazontitleemail{luhanqin@amazon.com}}
}

\begin{document}

\maketitle
\fancyhead{}
\renewcommand{\headrulewidth}{0pt}

\begin{abstract}
AI research agents automate the cycle of proposing, implementing, and evaluating experiments, opening a path toward recursive self-improvement. Yet their ability to propose experiments outpaces their capacity to execute them in real environments, making outcome prediction a key capability for sustained self-improvement under limited experimental budgets. We investigate language models as Research World Models (RWMs), which predict the outcomes of candidate interventions across research environments. Our evaluation draws on over 2,600 experimental records from nine research environments spanning pretraining, post-training, and inference, representing more than 171,000 H100 GPU-hours of experimentation. Research knowledge acquired from real experimental experience improves RWM predictions of unseen interventions within the same environment (Spearman +0.27), and can be reused \emph{across} environments. For example, using only pretraining experience from OLMo3, Marin, and Nanochat, an RWM reduces selection regret in the Qwen3 environment by 78\% compared with zero-experience setting. These benefits extend to multi-round Autoresearch under a fixed selection budget: RWMs with in-env and cross-env research knowledge increase the best gain achieved by 15.8\% and 11.6\%, respectively. Ablations across 13 language models used as RWMs show that adding research knowledge can improve intervention ranking more than changing models or increasing reasoning effort alone. These findings support language models as RWMs and motivate accumulating experimental data for future RWM training.
\end{abstract}

\section{Introduction}
\label{sec:introduction}

AI research agents~\citep{yamada2025aiscientistv2,schmidgall2025agentlaboratory} are automating the cycle of proposing, implementing, and evaluating experiments \citep{lu2024aiscientist,karpathy2026autoresearch}, opening a path toward recursive self-improvement. Yet their ability to propose experiments outpaces their capacity to execute them: a plausible, implementable intervention is cheap to generate but expensive to test~\citep{kaplan2020scaling,hoffmann2022computeoptimal}, and deciding whether it deserves training or evaluation budget still requires anticipating its actual effects. For example, changing a learning-rate schedule or training-data mixture~\citep{xie2023doremi} may improve performance with one model and budget but offer little benefit, or even hurt performance, with another. Under limited budgets, this uncertainty directly affects which interventions can be explored and how quickly effective improvements are found. Predicting outcomes before execution therefore has direct research value: it helps researchers and research agents compare candidate interventions and allocate resources to more promising experiments~\citep{snoek2012practicalbo,klein2017learningcurve}. This need spans pretraining, post-training, and inference, motivating a predictive capability that can serve multiple research environments by anticipating intervention outcomes in each of them.

\begin{figure}[t]
\centering
\includegraphics[width=0.95\linewidth]{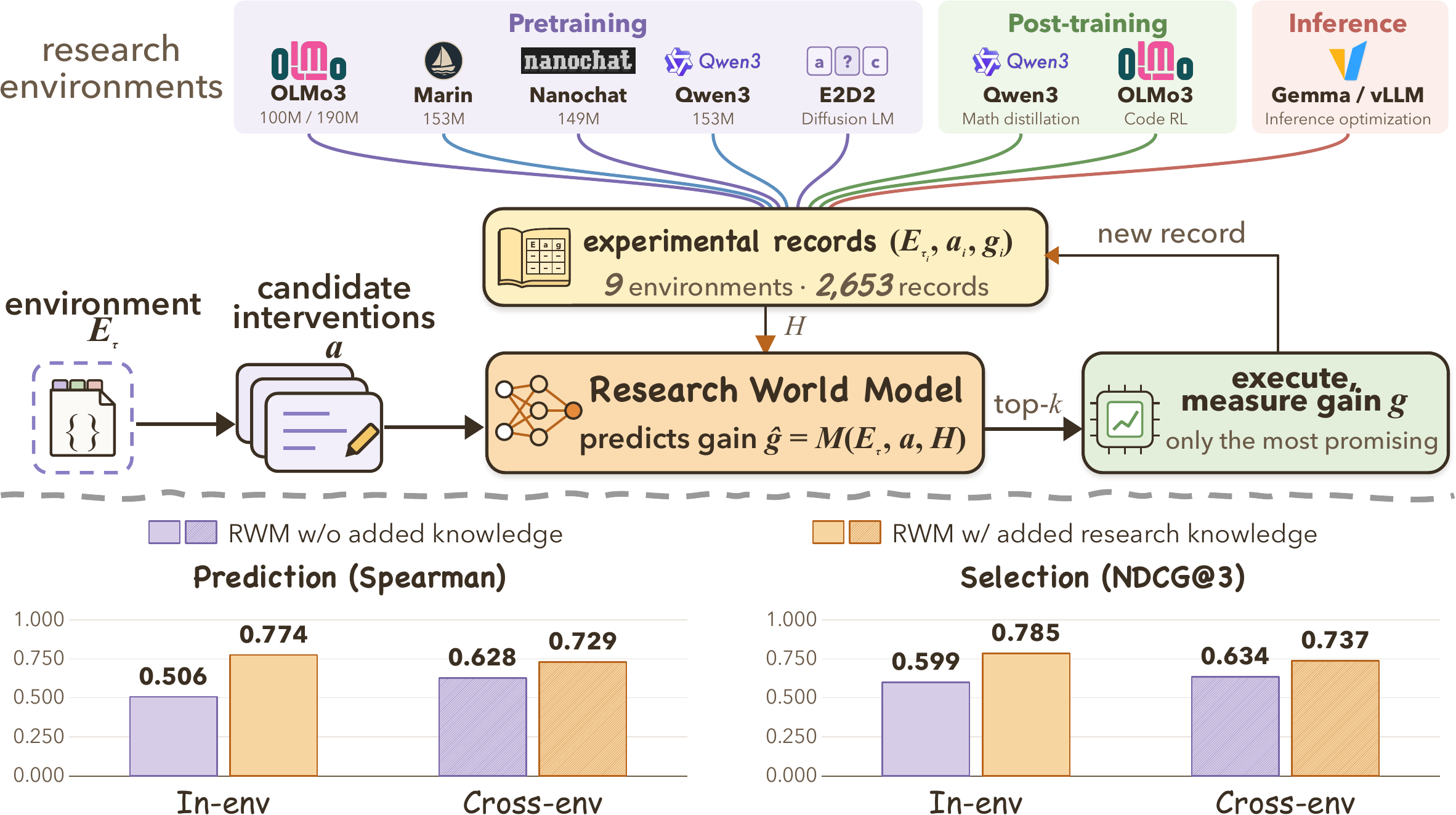}
\caption{\textbf{Large language models as Research World Models.} \textit{Top}: Given a research environment $E_\tau$ and candidate interventions $a$, an RWM predicts gains $\hat g$ before execution, so that only the most promising interventions consume experimental budget. Executed experiments become records $(E_{\tau_i},a_i,g_i)$ that form history $H$ for subsequent
predictions. Records may come from the same environment or from other environments, allowing the same RWM to serve a target environment that has no records of its own. \textit{Bottom}: Research knowledge improves experiment performance prediction and experiment selection, both within and across research environments. Bars show mean performance over the evaluated environments; see Tables~\ref{tab:result-summary}, \ref{tab:environment-holdout},
\ref{tab:seen-selection}, and~\ref{tab:transfer-selection} for detailed results.}
\label{fig:teaser}
\end{figure}

To satisfy such prediction needs, we formulate and study the Research World Model (RWM): a single model that serves multiple research environments by predicting the outcome of executing a candidate intervention in its respective environment (Figure~\ref{fig:teaser}). This perspective draws on world models~\citep{hafner2019planet,schrittwieser2020muzero,chua2018pets}, which support decisions by predicting the consequences of actions \citep{ha2018world,hafner2020dreamer}; here, the experimental intervention plays the role of the action, the research environment provides the context in which it is taken, and the experimental outcome is the consequence to be predicted. The research environment specifies the reference model, data, experimental configuration, budget, and evaluation setup. An experimental intervention specifies one or more executable changes relative to this starting point, and its outcome is measured under the environment's evaluation protocol after execution. Researchers or research agents can use RWM predictions to compare candidates, select the next experiment, and incorporate newly observed outcomes into subsequent decisions.

Language models are a natural candidate for implementing RWMs: they possess prior knowledge of models and training methods, and they can read descriptions of research environments and candidate interventions expressed in natural language, code, and configurations~\citep{brown2020language,chen2021codex}. Prior work has shown that language models can use experimental experience within a single task~\citep{zheng2026can} to assist optimization and research \citep{llambo2024,foster2026ai}. The interventions we consider range from hyperparameter adjustments to changes in model architectures, training objectives, and optimization mechanisms, applied in research environments with different foundation models, training recipes, and compute budgets. Because the same intervention can have different effects in different environments, predicting its outcome requires accounting for the environment in which it is applied. We therefore focus on a question that prior studies have not directly answered: \textit{can research knowledge acquired from real experiments in one environment help predict intervention outcomes in another?} This question concerns whether experimental experience can accumulate into research knowledge that serves multiple research environments, rather than remaining tied to the environment that produced it.

Our evaluation draws on 2,653 experimental records from nine research environments spanning pretraining, post-training, and inference, representing 171,862 H100 GPU-hours of real experimentation. We distinguish two prediction settings: in-environment prediction (in-env) uses experimental records from the same environment to predict unseen interventions, whereas cross-environment prediction (cross-env) uses records from other environments to predict unseen interventions in a target environment. Our experiment reveals that large language models can learn from real experimental experience: across five environments in the in-env setting, research knowledge raises mean Spearman correlation on unseen interventions from 0.506 to 0.774. This knowledge can also be reused across environments: without any target-environment records, knowledge from other environments raises mean Spearman correlation from 0.628 to 0.729 across five pretraining environments. These predictive gains translate into better experiment selection. When selecting three of sixteen candidates, in-env research knowledge raises mean normalized discounted cumulative gain at rank three (NDCG@3) from 0.599 to 0.785 across five environments, and cross-env knowledge reduces macro-average Regret@3 by 52.2\% across five autoregressive pretraining environments. In multi-round Autoresearch on OLMo3-100M, initializing RWMs with in-env or cross-env research knowledge increases mean final best gain by 15.8\% and 11.6\%, respectively, relative to no initial research knowledge under the same budget of 24 experiment selections (Figure~\ref{fig:replay-discovery}).

Ablations on OLMo3-100M further show that adding research knowledge improves intervention ranking more than changing models or increasing reasoning effort alone. Across 13 language models, Spearman correlations without additional research knowledge range from $-0.362$ to 0.621; with research knowledge they range from 0.638 to 0.896, so even the weakest model with knowledge exceeds the strongest model without it. Holding the model fixed, raising reasoning effort from LOW to MAX increases Spearman correlation from 0.609 only to 0.648, whereas keeping LOW effort and adding research knowledge raises it to 0.892. Within the tested settings, insufficient relevant research knowledge is thus a more prominent bottleneck for RWM prediction than limited reasoning effort, supporting the accumulation of research knowledge as a central direction for developing RWMs.

Together, these results establish the empirical viability of large language models as RWMs, connecting (1) learning from experimental experience, (2) reuse of research knowledge across environments, and (3) value for research decisions. They show that real experiments do more than test the interventions at hand: \textit{they produce research knowledge that benefits subsequent research, within and beyond the environment that generated it}. This work acquires and reuses such knowledge through in-context learning~\citep{garg2022incontext,xie2022implicitbayesian}, providing an empirical basis for collecting research data and training dedicated models. Our long-term goal is to train RWMs on large and diverse collections of research data, enabling them to continually accumulate research knowledge and develop predictive capabilities that serve multiple research environments, so that research agents can direct limited experimental budgets toward the most promising interventions.

\section{Language Models as Research World Models}
\label{sec:research-world-model}
\label{sec:rl-environment}

\subsection{Research Environments and Interventions}
\label{sec:research-environments}

\begin{samepage}
Each research task $\tau$ defines an environment $E_\tau$. The environment specifies a reference system and the conditions under which experiments are executed and evaluated. Let $\mathcal A_\tau$ denote the agent's admissible intervention space. Conditions outside this scope of control remain fixed, including those imposed by an experimental protocol.

\end{samepage}

Given a history $H$ of past experiments, the agent chooses a materialized intervention $a\in\mathcal A_\tau$: a concrete change relative to the reference code or recipe, possibly combining several edits. A complete experimental rollout produces a terminal state $s$. An evaluation function $g_\tau$ computes the gain $g$ from this state, defined as the change in benchmark score relative to the reference $r$:
\begin{equation}
s\sim E_\tau(\cdot\mid a),
\qquad g=g_\tau(s).
\label{eq:research-environment}
\end{equation}
Here $E_\tau(\cdot\mid a)$ denotes the distribution of terminal states, which contain experimental artifacts such as model checkpoints. For bits per byte (BPB), gain is the reference score minus the candidate score; for accuracy and throughput, gain is the candidate score minus the reference score. When accuracy scores are fractions in $[0,1]$, their difference is multiplied by 100 to report gain in percentage points (pp). Final performance gain is our primary outcome, with positive values indicating improvement.

We study a fixed-reference setting: experiments within each environment start from the same reference while $H$ accumulates. Cross-environment transfer tests whether experimental experience from other environments helps predict outcomes in a target environment, for example when the reference model or training implementation differs.

\subsection{LLMs as Research World Models}
\label{sec:rwm-prediction}

\textbf{A research world model (RWM)} approximates how $E_\tau$ responds to a candidate intervention before it is executed. Let $M$ denote a shared RWM that uses the available history $H$. We focus on predicting the final gain $g_\tau(s)$ computed from the terminal state $s$. The prediction $\hat g = M(E_\tau,a,H)$ targets the median gain under the rollout distribution in Equation~\ref{eq:research-environment}:
\begin{equation}
\hat g = M(E_\tau,a,H)\approx
\operatorname{Median}_{s\sim E_\tau(\cdot\mid a)}\!\left[g_\tau(s)\right].
\label{eq:rwm-definition}
\end{equation}
Conditioning on $\tau$ allows the same $M$ to assess interventions across environments. History $H$ informs prediction, and the rollout distribution $E_\tau(\cdot\mid a)$ depends on the task and intervention.

\textbf{LLMs-as-RWMs.} LLMs provide an implementation of $M$ by combining prior knowledge of models and training methods with the ability to interpret code, configurations, and experimental records. Given $(E_\tau,a,H)$, an LLM can draw on this knowledge and evidence to infer how a proposed intervention interacts with the target environment and affects the experimental outcome. This inference may be implicit in the model's computation or articulated through an explicit reasoning trace. We hypothesize that such inference approximates the rollout's consequences for the gain $g_\tau(s)$.
Let $\phi$ map the raw observations $(E_\tau,a,H)$ to an input textual representation ($\phi$ can be the identity map or a compression). With LLM parameters $\theta$ held fixed, we implement the RWM as
\begin{equation}
\hat g = M(E_\tau,a,H)
=\operatorname{LLM}_\theta\!\left(\phi(E_\tau,a,H)\right).
\label{eq:frozen-rwm-predictor}
\end{equation}
In our implementation, $\phi$ serializes the task description, materialized intervention, and permitted history as text. We instruct the LLM to return the median estimate in Equation~\ref{eq:rwm-definition}, expressed in the same units as $g_\tau(s)$. Our experiments assess this approximation through prediction accuracy, transfer across environments, and its value for experiment selection.

\subsection{Learning from and Transferring Experimental Experience}
\label{sec:rwm-experience}
\label{sec:cross-environment-rwm}

To study the learnability and transferability of LLMs-as-RWMs, we formulate research experience across environments as triplets.
For experiment $i$, let $\tau_i$, $a_i$, $s_i$, and $g_i$ denote its task, executed intervention, terminal state, and observed gain. We write a history of $n$ experiments as
\begin{equation}
\begin{aligned}
H&=\bigl((E_{\tau_i},a_i,g_i)\bigr)_{i=1}^{n},\\
s_i&\sim E_{\tau_i}(\cdot\mid a_i),
\qquad g_i=g_{\tau_i}(s_i).
\end{aligned}
\label{eq:rwm-history}
\end{equation}
Each record also retains reference information, measurements, and execution metadata. Let $\tau_*$ denote the target task and $E_{\tau_*}$ its environment. The history $H$ contains $n=|H|$ records, with $E_{\tau_i}$ denoting the environment of its $i$-th record. We deterministically serialize $H$ as part of $\phi(E_{\tau_*},a,H)$, providing experimental experience for in-context learning \citep{brown2020language}.
We consider three history conditions:
\begin{equation}
\begin{array}{@{}l@{\qquad}l@{\qquad}l@{}}
\text{No experimental history }(W_0) & n=0 & \\[3pt]
\text{In-environment }(W_{\mathrm{in-env}}) & n>0 & E_{\tau_i}=E_{\tau_*} \quad \forall i,\\[3pt]
\text{Cross-environment }(W_{\mathrm{cross-env}}) & n>0 & E_{\tau_i}\neq E_{\tau_*} \quad \forall i.
\end{array}
\label{eq:rwm-history-conditions}
\end{equation}
In all three conditions, the model receives the target task $\tau_*$ and queried intervention $a$. Gains in source records are measured against their own tasks' references. For the cross-environment evaluation, source histories exclude each held-out intervention and its near-duplicate group (Section~\ref{sec:unseen-environments}). The queried intervention's outcome and post-execution metadata are withheld in every condition. Appendices~\ref{app:record-schema} and~\ref{app:prediction-inputs} provide the record schema and prediction prompt template.

\section{Experiments}
\label{sec:experiments}

\begin{table}[t]
\centering
\footnotesize
\setlength{\tabcolsep}{2.5pt}
\renewcommand{\arraystretch}{1.12}
\caption{\textbf{Research environments and experimental scale.} In-env and Cross-env indicate prediction-evaluation coverage.}
\label{tab:overall-experiment-matrix}
\begin{tabularx}{\linewidth}{@{}>{\raggedright\arraybackslash}p{.18\linewidth}>{\raggedright\arraybackslash}p{.20\linewidth}>{\raggedright\arraybackslash}Xrrcc@{}}
\toprule
\shortstack[l]{Research Env.} & \shortstack[l]{Model /\ System} & \shortstack[l]{Eval metric} & Records & \shortstack{H100\\GPU-hours} & In-env & Cross-env \\
\midrule
\multicolumn{7}{@{}l}{\textbf{Pretraining}} \\
OLMo3-100M  & OLMo3-100M & mean BPB $\downarrow$ & 480 & 8,572 & $\checkmark$ & $\checkmark$ \\
Marin & Marin-153M & mean BPB $\downarrow$ & 132 & 12,101 & --- & $\checkmark$ \\
Nanochat & Nanochat-149M & mean BPB $\downarrow$ & 132 & 2,584 & --- & $\checkmark$ \\
Qwen3 & Qwen3-153M & mean BPB $\downarrow$ & 132 & 15,447 & --- & $\checkmark$ \\
OLMo3-190M & OLMo3-190M & mean BPB $\downarrow$ & 132 & 5,582 & --- & $\checkmark$ \\
Diffusion & E2D2-170M & DUEL BPB $\downarrow$ & 661 & 47,592 & $\checkmark$ & --- \\
\midrule
\multicolumn{7}{@{}l}{\textbf{Post-training}} \\
Math distillation & Qwen3-1.7B & AIME24/25 avg@32 $\uparrow$ & 280 & 10,432 & $\checkmark$ & --- \\
Code RL & OLMo-3-7B & LiveCodeBench~$\uparrow$ & 149 & 60,672 & $\checkmark$ & --- \\
\midrule
\multicolumn{7}{@{}l}{\textbf{Inference}} \\
Inference optimization & Gemma-4-E4B-it / vLLM & Quality-constrained throughput (tokens/s) $\uparrow$ & 555 & 8,880 & $\checkmark$ & --- \\
\midrule
\multicolumn{3}{@{}l}{\textbf{Total}} & \textbf{2,653} & \textbf{171,862} & \textbf{5} & \textbf{5} \\
\bottomrule
\end{tabularx}
\end{table}

\subsection{Experimental Setup}
\label{sec:experiments-setup}

\paragraph{Research environments and evaluation overview.}

Our evaluation spans nine research environments across pretraining, post-training, and inference (Table~\ref{tab:overall-experiment-matrix}). The corpus totals 2,653 records, accounting for 171,862 H100 GPU-hours. We first evaluate whether research knowledge improves outcome prediction within and across environments. We then test whether these predictive gains translate into better experiment selection under a fixed budget and better outcomes in multi-round Autoresearch. Finally, we compare the predictive benefits of research knowledge, backbone choice, and reasoning effort.

\paragraph{RWM settings and baselines.}

By default, we use Claude Opus 5 as the LLM backbone of the RWM, with a 1M context window and HIGH reasoning effort. We use $W_0$ as a baseline to assess the contribution of experimental records. When comparing knowledge conditions, we hold the RWM backbone and inference settings fixed. We include $k$-nearest-neighbor regression (kNN) \citep{altman1992knn} and ridge regression \citep{hoerl1970ridge}, which receive the same experimental records and candidate inputs as the record-conditioned RWM. For each prediction evaluation, we fix the history--test split and evaluate all methods on the same test set. Appendix~\ref{app:baseline-protocols} specifies the baseline inputs and hyperparameter selection.

\FloatBarrier

\begin{table}[t]
\centering
\footnotesize
\setlength{\tabcolsep}{3pt}
\renewcommand{\arraystretch}{1.18}
\caption{\textbf{Within-environment prediction performance.} Each environment uses a single fixed split. History denotes context records supplied to $W_{\mathrm{in-env}}$; Test denotes the number of held-out candidates.}
\label{tab:result-summary}
\begin{tabularx}{\linewidth}{@{}>{\raggedright\arraybackslash}Xrrrrr>{\columncolor{rwmshade}}rrrr>{\columncolor{rwmshade}[\tabcolsep][0pt]}r@{}}
\toprule
& & & \multicolumn{4}{c}{MAE $\downarrow$} & \multicolumn{4}{c}{Spearman $\uparrow$} \\
\cmidrule(lr){4-7}\cmidrule(lr){8-11}
Environment (unit) & History & Test & $W_0$ & kNN & ridge & $W_{\mathrm{in-env}}$ & $W_0$ & kNN & ridge & $W_{\mathrm{in-env}}$ \\
\midrule
OLMo3-100M (BPB) & 320 & 160 & 2.93e-2 & 2.58e-2 & 3.13e-2 & \textbf{2.03e-2} & 0.62 & 0.78 & 0.67 & \textbf{0.90} \\
Diffusion (BPB) & 536 & 125 & 1.95e-2 & 1.58e-2 & 1.99e-2 & \textbf{8.77e-3} & 0.46 & 0.49 & 0.50 & \textbf{0.78} \\
Math distillation (pp) & 180 & 100 & 1.58 & 1.15 & 1.11 & \textbf{0.91} & 0.43 & 0.61 & 0.62 & \textbf{0.70} \\
Code RL (pp) & 34 & 115 & 5.98 & 5.45 & 5.02 & \textbf{4.17} & 0.48 & 0.29 & 0.48 & \textbf{0.62} \\
Inference optimization (tokens/s) & 457 & 98 & 13.19 & 13.18 & 12.73 & \textbf{1.34} & 0.54 & 0.62 & 0.72 & \textbf{0.88} \\
\bottomrule
\end{tabularx}
\end{table}
\begin{table}[t]
\centering
\footnotesize
\setlength{\tabcolsep}{3pt}
\renewcommand{\arraystretch}{1.15}
\caption{\textbf{Cross-environment prediction performance.} Macro is the equal-weight mean over target environments.}
\label{tab:environment-holdout}
\begin{tabularx}{\linewidth}{@{}>{\raggedright\arraybackslash}Xrrr>{\columncolor{rwmshade}}rrrr>{\columncolor{rwmshade}[\tabcolsep][0pt]}r@{}}
\toprule
& \multicolumn{4}{c}{MAE (BPB) $\downarrow$} & \multicolumn{4}{c}{Spearman $\uparrow$} \\
\cmidrule(lr){2-5}\cmidrule(lr){6-9}
\shortstack{Target\\Environment} & $W_0$ & kNN & ridge & $W_{\mathrm{cross-env}}$ & $W_0$ & kNN & ridge & $W_{\mathrm{cross-env}}$ \\
\midrule
OLMo3-100M & 1.11e-2 & 1.18e-2 & 1.65e-2 & \textbf{9.28e-3} & 0.69 & 0.56 & 0.26 & \textbf{0.83} \\
OLMo3-190M & \textbf{7.94e-3} & 1.08e-2 & 1.19e-2 & 8.56e-3 & 0.72 & 0.43 & 0.26 & \textbf{0.78} \\
Qwen3 & 9.20e-3 & 1.13e-2 & 1.69e-2 & \textbf{7.80e-3} & 0.74 & 0.71 & 0.40 & \textbf{0.82} \\
Marin & 9.35e-3 & 1.13e-2 & 1.03e-2 & \textbf{8.18e-3} & 0.42 & 0.28 & 0.43 & \textbf{0.50} \\
Nanochat & 2.79e-2 & 3.61e-2 & 3.57e-2 & \textbf{2.70e-2} & 0.57 & $-0.17$ & $-0.12$ & \textbf{0.71} \\
\midrule
Macro & 1.31e-2 & 1.63e-2 & 1.82e-2 & \textbf{1.22e-2} & 0.63 & 0.36 & 0.25 & \textbf{0.73} \\
\bottomrule
\end{tabularx}
\end{table}

\subsection{Research Knowledge Improves RWM Predictions}
\label{sec:prediction}

\paragraph{Prediction settings and metrics.}

We evaluate predictions of unseen interventions in two settings: in-environment (in-env), where records come from the evaluated environment, and cross-environment (cross-env), where they come from other environments. We use mean absolute error (MAE) to measure the difference between predicted and measured gains, where lower values are better, and Spearman correlation to measure agreement between rankings by predicted and measured gains, where higher values are better.

\subsubsection{In-Environment Prediction (In-Env)}
\label{sec:seen-environments}
\label{sec:seen-prediction}

\paragraph{Experimental records improve in-environment RWM predictions.}
Compared with $W_0$, $W_{\mathrm{in-env}}$ improves gain prediction and intervention rankings in all five environments, with a mean Spearman increase of 0.27 (Table~\ref{tab:result-summary}). The record-conditioned RWM also outperforms kNN and ridge on both metrics in every environment, which shows that language models acting as RWMs can learn more effectively from experimental experience.

\subsubsection{Cross-Environment Prediction (Cross-Env)}
\label{sec:unseen-environments}

\paragraph{Cross-environment evaluation.}

We take each of five autoregressive pretraining environments in turn as the target. $W_{\mathrm{cross-env}}$ uses 244 records from the other four environments to predict gains for 71 interventions in the target environment. The target environment and reference setting are provided, with no target-environment records. The held-out interventions and their near-duplicate groups are excluded from the source records. kNN and ridge receive the same source records.

\begin{table}[t]
\centering
\footnotesize
\setlength{\tabcolsep}{2pt}
\renewcommand{\arraystretch}{1.18}
\caption{\textbf{Mean within-environment experiment-selection performance} (selecting 3 of 16 candidates per pool). Metrics are averaged over 200 shared random partitions of each environment's fixed test set.}
\label{tab:seen-selection}
\begin{tabularx}{\linewidth}{@{}>{\raggedright\arraybackslash}Xrrrr>{\columncolor{rwmshade}}rrrrr>{\columncolor{rwmshade}[\tabcolsep][0pt]}r@{}}
\toprule
& \multicolumn{5}{c}{Regret@3 $\downarrow$} & \multicolumn{5}{c}{NDCG@3 $\uparrow$} \\
\cmidrule(lr){2-6}\cmidrule(lr){7-11}
Environment (unit) & Random & $W_0$ & kNN & ridge & $W_{\mathrm{in-env}}$ & Random & $W_0$ & kNN & ridge & $W_{\mathrm{in-env}}$ \\
\midrule
OLMo3-100M (BPB) & 1.32e-2 & 2.80e-3 & 1.81e-3 & 2.14e-3 & \textbf{7.00e-4} & 0.28 & 0.70 & 0.78 & 0.72 & \textbf{0.84} \\
Diffusion (BPB) & 1.09e-2 & 3.14e-3 & 1.79e-3 & 5.47e-3 & \textbf{4.70e-4} & 0.16 & 0.51 & 0.62 & 0.32 & \textbf{0.87} \\
Math distillation (pp) & 0.87 & 0.48 & 0.41 & 0.45 & \textbf{0.29} & 0.20 & 0.33 & 0.38 & 0.34 & \textbf{0.47} \\
Code RL (pp) & 4.16 & 2.20 & 2.12 & 1.54 & \textbf{0.83} & 0.42 & 0.59 & 0.64 & 0.73 & \textbf{0.81} \\
Inference optimization (tokens/s) & 4.31 & 3.60 & \textbf{1.62} & 2.09 & 2.53 & 0.69 & 0.86 & 0.93 & 0.92 & \textbf{0.93} \\
\bottomrule
\end{tabularx}
\end{table}
\begin{table}[t]
\centering
\footnotesize
\setlength{\tabcolsep}{2pt}
\renewcommand{\arraystretch}{1.18}
\caption{\textbf{Mean cross-environment experiment-selection performance} (selecting 3 of 16 candidates per pool). Metrics are averaged over 200 shared random partitions of each target environment's fixed test set. Macro is the equal-weight mean over target environments.}
\label{tab:transfer-selection}
\begin{tabularx}{\linewidth}{@{}>{\raggedright\arraybackslash}Xrrrr>{\columncolor{rwmshade}}rrrrr>{\columncolor{rwmshade}[\tabcolsep][0pt]}r@{}}
\toprule
& \multicolumn{5}{c}{Regret@3 (BPB) $\downarrow$} & \multicolumn{5}{c}{NDCG@3 $\uparrow$} \\
\cmidrule(lr){2-6}\cmidrule(lr){7-11}
Target & Random & $W_0$ & kNN & ridge & $W_{\mathrm{cross-env}}$ & Random & $W_0$ & kNN & ridge & $W_{\mathrm{cross-env}}$ \\
\midrule
OLMo3-100M & 1.31e-2 & 4.27e-3 & 4.74e-3 & 4.84e-3 & \textbf{2.24e-3} & 0.26 & 0.65 & 0.66 & 0.66 & \textbf{0.84} \\
OLMo3-190M & 8.42e-3 & 1.64e-3 & 1.60e-3 & 1.96e-3 & \textbf{5.50e-4} & 0.24 & 0.68 & 0.66 & 0.64 & \textbf{0.81} \\
Qwen3 & 1.51e-2 & 9.40e-4 & 3.50e-4 & 1.72e-3 & \textbf{2.10e-4} & 0.27 & 0.77 & 0.85 & 0.73 & \textbf{0.91} \\
Marin & 1.07e-2 & \textbf{7.40e-4} & 1.10e-3 & 8.40e-4 & 7.90e-4 & 0.18 & \textbf{0.74} & 0.61 & 0.67 & 0.70 \\
Nanochat & 8.27e-3 & 4.94e-3 & 1.53e-2 & 1.51e-2 & \textbf{2.22e-3} & 0.14 & 0.32 & 0.12 & 0.06 & \textbf{0.43} \\
\midrule
Macro & 1.11e-2 & 2.51e-3 & 4.62e-3 & 4.89e-3 & \textbf{1.20e-3} & 0.22 & 0.63 & 0.58 & 0.55 & \textbf{0.74} \\
\bottomrule
\end{tabularx}
\end{table}

\paragraph{Experimental records support cross-environment RWM predictions.}

$W_{\mathrm{cross-env}}$ achieves higher Spearman rank correlation than $W_0$ in all five target environments, with an average increase of 0.10 (Table~\ref{tab:environment-holdout}); paired bootstrap confidence intervals are reported in Appendix~\ref{app:cross-env-uncertainty}. Yet kNN, which achieves higher Spearman correlation than $W_0$ in four of the five in-env environments, falls below it on every transfer target (macro-average Spearman 0.36 vs. 0.63). kNN can struggle to capture how interventions depend on their environment: semantically similar changes may correspond to different modifications under different target recipes.

For example, Nanochat's reference setting already uses Muon. For a Muon optimizer variant, kNN predicts a gain of $+0.018$ BPB based on source-environment records of similar interventions, most of which show positive gains. The measured gain on Nanochat is $-0.036$ BPB, while $W_0$ predicts $-0.035$ BPB. This case illustrates how records from source environments can mislead target-env predictions when reference settings differ across environments.

\subsection{RWM Predictions Improve Experiment Selection}
\label{sec:experiment-selection}

\paragraph{Budget allocation settings and metrics.}

While good global rankings (Spearman) do not guarantee that the few highest-gain candidates are identified, an agent in a typical Autoresearch round must choose the most promising few proposals to execute. To evaluate RWM predictions in this setting, we use an execution budget of roughly 20\% by selecting 3 interventions from each pool of 16 candidates. Specifically, we evaluate 200 random partitions of the test set into shared pools and average the evaluation results. \textit{Random} selects 3 candidates uniformly at random from each pool.

Regret@3 measures the gain lost by missing the best candidate. NDCG@3 evaluates the positive gains and ordering of the selected interventions. Appendix~\ref{app:selection-protocol} defines the metrics, candidate grouping, and aggregation.

\paragraph{Research knowledge improves experiment selection with RWMs.}

In in-env selection, $W_{\mathrm{in-env}}$ achieves an average relative reduction of 58.2\% in Regret@3 compared with $W_0$ across the five environments. NDCG@3 also improves in every environment (Table~\ref{tab:seen-selection}).

In cross-env selection, $W_{\mathrm{cross-env}}$ reduces average Regret@3 by 52.2\% relative to $W_0$. Both Regret@3 and NDCG@3 improve in all targets except Marin (Table~\ref{tab:transfer-selection}). Both regression-based baselines have higher average Regret@3 than $W_0$. Together, these results show that RWMs can use research knowledge from both the same and other research environments to select higher-gain interventions under the same execution budget (see Appendix~\ref{app:rwm-prediction-cost} for RWM prediction costs).

\subsection{Multi-Round Autoresearch}
\label{sec:seen-sequential}
\label{sec:replay}

\paragraph{Multi-round Autoresearch settings.}

We further examine whether RWMs can discover higher-gain interventions under the same experiment-selection budget in multi-round Autoresearch. We conduct five Autoresearch evaluations on OLMo3-100M, each with eight rounds. Sixteen new candidate interventions are introduced each round, and each method selects three interventions from those currently available, for a total selection budget of 24 per evaluation. We compare three initial RWM settings: $W_0$ receives no experimental records, $W_{\mathrm{in-env}}$ receives 244 records from the same environment, and $W_{\mathrm{cross-env}}$ receives 244 records from four other different environments. We also compare against kNN and ridge using the same cross-environment records. Within each evaluation, all methods receive the same new candidates each round and have the same selection budget. Results are averaged over the five evaluations. Appendix~\ref{app:sequential-allocation} specifies candidate arrivals, experimental feedback, and scoring.

\begin{figure}[t]
\centering
\includegraphics[width=0.9\linewidth]{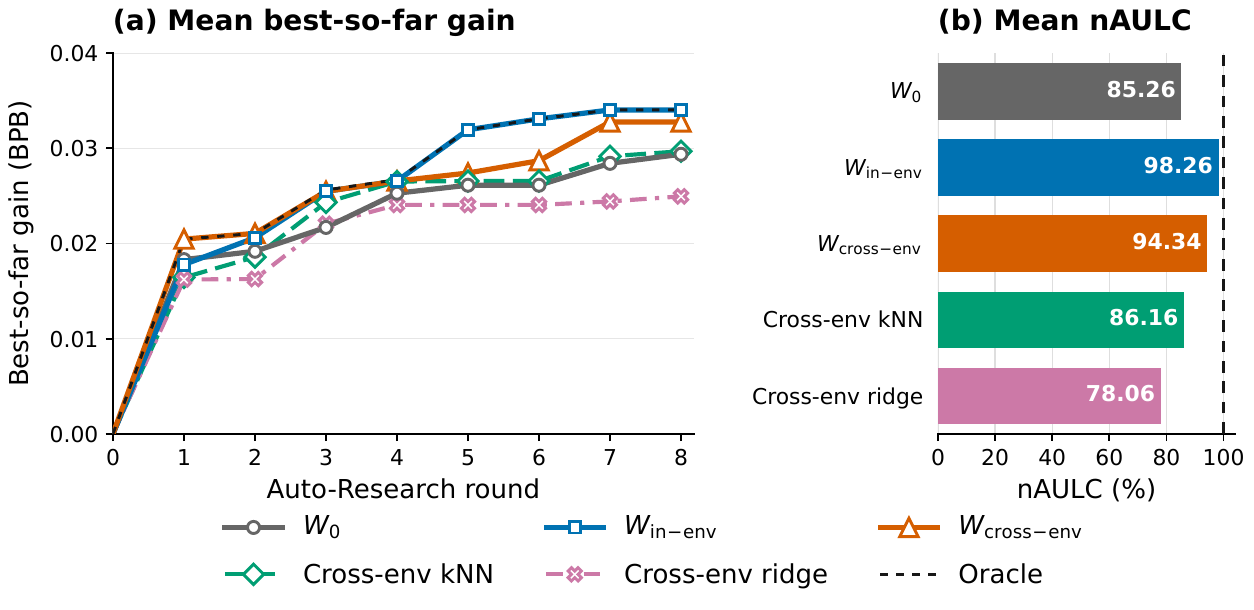}
\caption{\textbf{Multi-round Autoresearch} on OLMo3-100M. Means over five Autoresearch evaluations, each with eight rounds. \textbf{(a)} Best-so-far gain (BPB). \textbf{(b)} nAULC (\%), computed by normalizing each evaluation's curve area by that of its matched Arrival Oracle and then averaging across evaluations.}
\label{fig:replay-discovery}
\end{figure}

\paragraph{Evaluation metrics.}

We use best-so-far gain to track the highest measured gain among the interventions selected up to each round, with the unchanged reference providing a zero-gain option. Its value after the final round is the final best gain. We use normalized area under the learning curve (nAULC) to assess the entire research process: finding high-gain interventions earlier yields a higher nAULC. The 100\% reference for nAULC is an ideal selector (Arrival Oracle) that knows the measured gains in advance but receives the same new candidates each round and has the same selection budget as the other methods. Higher values are better for both metrics.

\paragraph{Research knowledge improves multi-round outcomes with RWMs.}

Both in-environment and cross-environment records improve early discovery and final gains with RWMs. Initializing RWMs with in-environment or cross-environment records increases mean final best gain by 15.8\% and 11.6\%, respectively, relative to $W_0$ under the same budget of 24 selections (Figure~\ref{fig:replay-discovery}a). Cross-environment records approach the performance of within-environment records and outperform the regression-based baselines initialized with the same source records (Figure~\ref{fig:replay-discovery}b). These results extend the value of cross-environment knowledge reuse from single-round experiment selection to multi-round research.

\FloatBarrier

\subsection{Research Knowledge, Backbone Choice, and Reasoning Effort}
\label{sec:records-vs-model}

\paragraph{Comparison settings.}

In the OLMo3-100M environment, we compare 13 RWM backbones and five reasoning-effort levels for Claude Opus 5. Comparisons of LLM backbones and reasoning effort use the same test set and history records. Each configuration is evaluated with and without experimental records using Spearman correlation. Appendix~\ref{app:model-comparisons} gives the comparison settings.

\begin{figure}[t]
\centering
\includegraphics[width=0.9\linewidth]{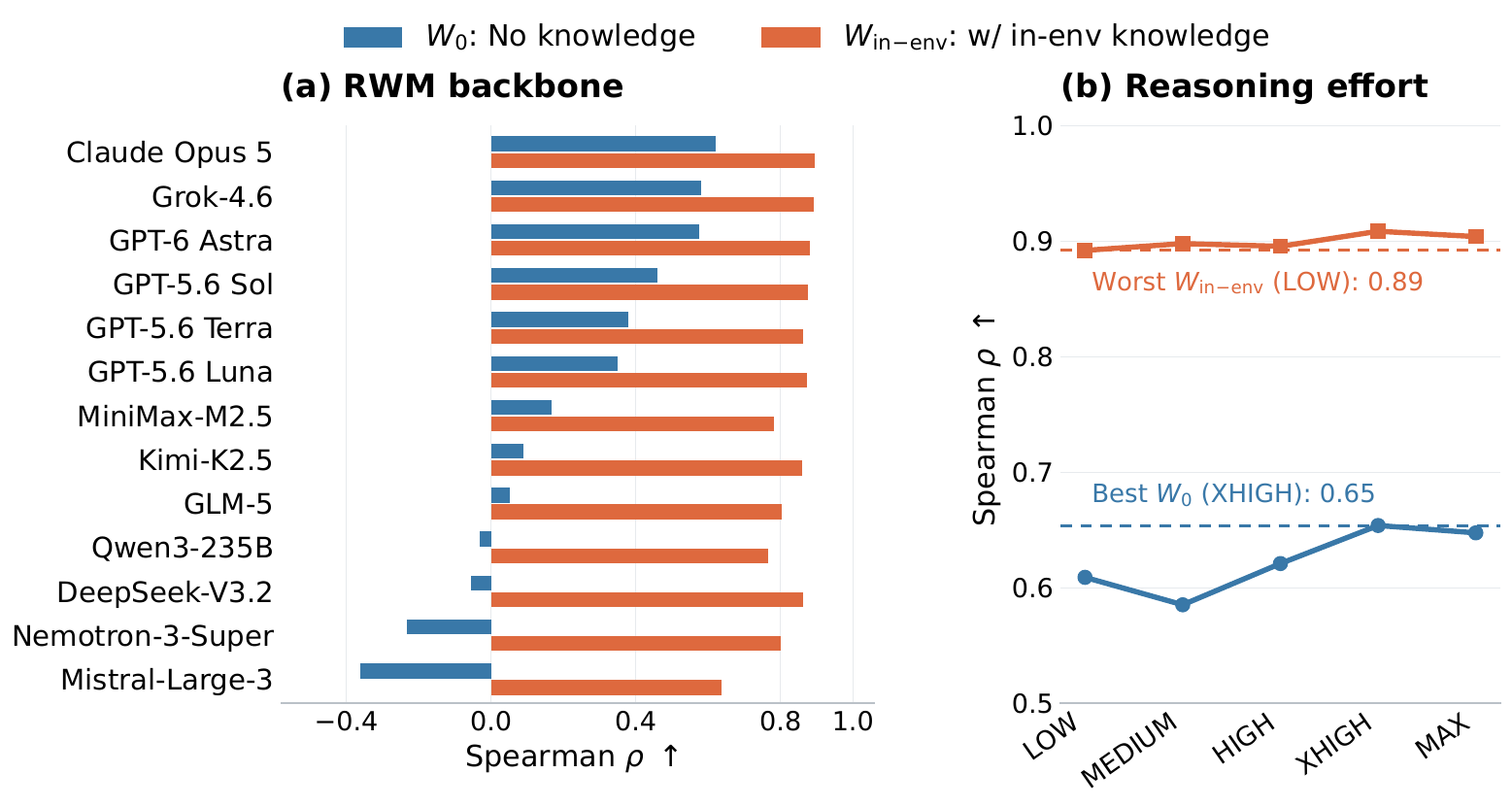}
\caption{RWM prediction performance on OLMo3-100M across \textbf{(a)} backbones and \textbf{(b)} Claude Opus 5 reasoning-effort settings.}
\label{fig:model-effort-ablations}
\end{figure}

\paragraph{Research knowledge has a larger effect than reasoning effort.}

Providing experimental records improves RWM prediction performance more than increasing reasoning effort: even LOW with records outperforms the best-performing effort setting without records (0.89 vs. 0.65; Figure~\ref{fig:model-effort-ablations}b).

\paragraph{The benefit extends across RWM backbones.}

Experimental records improve rankings for all 13 backbones, and the two performance ranges do not overlap: the lowest-performing backbone (Mistral-Large-3, 0.64) with records exceeds the highest-performing backbone (Claude Opus 5, 0.62) without them (Figure~\ref{fig:model-effort-ablations}a). In this environment, changing the backbone alone does not compensate for missing experimental evidence. Together, these two findings motivate collecting more experimental records to expand the research knowledge available to LLM-based RWMs and further improve their predictions.

\section{Related Work}
\label{sec:related-work}

\paragraph{Automated research agents.}
Automated research agents organize AI research as an iterative process
of proposing, implementing, and evaluating experiments.
Autoresearch, AIRA$_2$, and the AI Scientist instantiate this process
with different search and execution workflows
\citep{karpathy2026autoresearch,aira2026,lu2024aiscientist}.
AutoLLMResearch further learns experiment-configuration policies
through interaction with a multi-fidelity environment
\citep{autollmresearch2026}.
Although these systems automate substantial parts of research,
evaluating candidate experiments still consumes time and computational
resources.
Our work studies a complementary capability: predicting experimental
outcomes to help research agents decide which candidates merit
execution under limited budgets.

\paragraph{Predictive models for research decisions.}
Several approaches use language models to assess candidates before
full execution.
LLAMBO uses historical evaluations as in-context examples for
surrogate modeling and candidate sampling, producing numerical
predictions with empirical uncertainty estimates \citep{llambo2024}.
Research preference models compare candidate plans and code using
information from previously executed solutions \citep{foster2026ai}.
Predict-Before-Executing studies pairwise solution preference and
incorporates prediction into the ForeAgent predict-then-verify loop
\citep{zheng2026can}.
These works connect predictive evaluation with optimization and
research search, providing close precedents for using forecasts
to allocate experimental effort.

\paragraph{World models for research.}
World models approximate environment dynamics, enabling agents to
anticipate the consequences of actions and use these predictions
for planning and control
\citep{ha2018world,hafner2020dreamer}.
WMRL applies this perspective to research-agent training, using
language-model predictions of execution outcomes as surrogate
rewards for RL post-training \citep{yang2026wmrl}.
We focus on whether experimental knowledge improves
intervention-outcome prediction within and across research environments,
including transfer without target-environment outcome records.
We further evaluate whether improved forecasts translate into better
experiment selection under fixed budgets.
Appendix~\ref{app:extended-related-work} discusses additional connections
to scientific forecasting and learning from experience.

\section{Discussion and Future Work}

Our results suggest that experiments produce reusable knowledge beyond testing individual interventions. Experimental records can improve prediction and guide subsequent research, including across environments. This motivates preserving environment descriptions, interventions, and outcomes, including experiments that do not improve performance. The demonstrated benefits arise through in-context learning, with cross-environment evidence currently limited to autoregressive pretraining.

A natural next step is to train RWMs on diverse experimental records and compare their generalization with in-context learning in unseen environments. Integrating RWMs into sustained research also raises the question of how to balance immediate improvement against acquiring knowledge for future decisions. Evaluating such systems requires measuring research progress and total costs, including data acquisition, inference, and execution, since better predictions alone do not guarantee more effective research.

\subsection*{AI use statement}

In this work, we used generative AI tools to refine hypotheses and provide feedback on the experimental design, to implement and debug the data-processing, evaluation, and analysis code, to clean and reformat experimental records into the tables reported in the paper, and to assist in interpreting results. We have not used generative AI tools to generate synthetic datasets, to develop the conceptual framework or notation, to formulate mathematical claims or write proofs, or for translation; qualitative and thematic data analysis is not applicable to this work. Additionally, we used generative AI tools to draft and edit parts of the paper, suggest its structure, summarize and identify related literature, format references, create or modify figures, and edit LaTex source. Separately, language models are the object of study in this paper and serve as the research world model under evaluation; that use is part of the method and is described in Sections~\ref{sec:research-world-model} and~\ref{sec:experiments} rather than here. We have reviewed all AI-assisted work: AI-generated code was inspected and tested by the authors, all reported numbers were recomputed from the raw experimental records, factual claims and citations were checked against source material, and the framing, methodology, and conclusions were decided by the authors. We take responsibility for the final content of this work, including text, claims, or artifacts produced with the aid of generative AI.

\subsection*{Ethics statement}

This work studies whether prior experimental records can reduce unnecessary experimental runs, but constructing those records and evaluating research world models also consumes substantial compute; Table~\ref{tab:overall-experiment-matrix} reports the H100 GPU-hours used to produce the experimental records so that efficiency claims can be weighed against this cost. The methods are dual use: better prediction and allocation could accelerate safety and efficiency research, but could equally accelerate capability development or concentrate advantages among organizations with large experimental archives. All experimental records were produced by the authors' own experimental runs on public models and datasets; the study involves no human subjects and collects no personally identifiable information. Before release, we will review records, code, and prompts for licensing and security concerns and document their provenance and known limitations.

\subsection*{Reproducibility statement}

The experimental records that all predictors consume are described in Appendix~\ref{app:record-schema}, and the research environments and benchmark scoring are given in Appendix~\ref{app:environments}. Appendix~\ref{app:prediction-inputs} provides a condensed prediction prompt template and record-context construction. Appendix~\ref{app:baseline-protocols} specifies the baseline inputs and hyperparameter selection, and Appendix~\ref{app:model-comparisons} describes the RWM-backbone and reasoning-effort comparisons. Appendix~\ref{app:static-evaluation} reports prediction confidence intervals and specifies the fixed-budget selection metrics and aggregation protocol, and Appendix~\ref{app:sequential-allocation} details candidate sampling, arrival schedules, and metric definitions for the multi-round protocol. Language-model predictions depend on a commercial API and are not bitwise reproducible; we fix the model version, context window, and reasoning effort (Section~\ref{sec:experiments-setup}), and report averages over repeated evaluations where sampling is involved.

\appendix
\section{Extended Related Work}
\label{app:extended-related-work}

\paragraph{Automated research agents.}
Automated research agents connect idea generation, implementation, execution, and feedback into iterative scientific workflows. ResearchAgent generates and revises research ideas using scientific literature and feedback from reviewing agents \citep{baek2024researchagent}; the AI co-scientist develops hypotheses through a generate--debate--evolve process guided by research objectives \citep{gottweis2025coscientist}. These systems address the construction and refinement of research proposals. The AI Scientist extends automation through experimentation and paper writing \citep{lu2024aiscientist}, while its second version introduces agentic tree search for experimental exploration \citep{yamada2025aiscientistv2}. Agent Laboratory organizes literature review, experimentation, and report writing around a human-provided research idea, with opportunities for human feedback \citep{schmidgall2025agentlaboratory}. Autoresearch iterates over code modifications, short training runs, and measured improvements \citep{karpathy2026autoresearch}. AIRA$_2$ scales experimental search through asynchronous execution, consistent evaluation, and interactive debugging \citep{aira2026}. AutoLLMResearch learns configuration policies in a multi-fidelity environment, including extrapolation from cheaper experiments to expensive configurations \citep{autollmresearch2026}. MLAgentBench and MLE-bench complement these systems with evaluations of multi-step machine learning experimentation and engineering \citep{huang2023mlagentbench,chan2024mlebench}. Together, these works establish the workflows and evaluation settings in which research prediction can be useful. When candidates outnumber affordable executions, forecasting offers a way to prioritize the experiments that receive resources.

\paragraph{Predictive models for research decisions.}
Predicting experimental performance to allocate computation has a substantial history. Bayesian optimization uses surrogate models and previous evaluations to select promising hyperparameters, including settings with variable evaluation costs \citep{snoek2012practicalbo}. Learning-curve models predict performance across configurations and can extrapolate partially observed training curves \citep{klein2017learningcurve}. Performance predictors for architecture search also use architecture, hyperparameter, and partial-training features to estimate final performance and enable early stopping \citep{baker2017performanceprediction}. These approaches illustrate several prediction interfaces; some use measurements from the candidate's ongoing execution, whereas our queried interventions are assessed without their execution outcomes. At a broader scale, language-model scaling laws relate loss to model size, data, and compute \citep{kaplan2020scaling}, compute-optimal training studies their allocation under a fixed budget \citep{hoffmann2022computeoptimal}, and DoReMi optimizes data-mixture weights with a proxy model before training a larger model \citep{xie2023doremi}. These findings provide concrete examples of why experimental conditions matter, without resolving prediction for arbitrary code-level interventions.
LLAMBO brings language models into surrogate modeling and candidate sampling through in-context historical evaluations, including numerical forecasts with empirical uncertainty estimates \citep{llambo2024}. Research preference models compare candidate plans and code with previously evaluated solutions; their inference-only and agentic variants differ in whether they additionally run pilot experiments \citep{foster2026ai}. Predict-Before-Executing studies pairwise solution preference, motivates prediction through implicit world modeling, and incorporates it into the ForeAgent predict-then-verify loop \citep{zheng2026can}. Further work compares research ideas by their expected empirical benchmark performance \citep{wen2025predicting,forecast2026}. Beyond AI research, SciPredict evaluates natural-science outcomes and reports limitations in accuracy and confidence calibration \citep{scipredict2026}; social-science forecasting by simulated participant responses achieves substantial correlations with measured treatment effects but systematically overestimates effect sizes \citep{ashokkumar2026}. These are precedents for forecasting and decision support, with reliability and information requirements that depend on the prediction task.

\paragraph{Experimental experience and in-context learning.}
In-context learning enables frozen language models to condition predictions on examples \citep{brown2020language}. Controlled studies show that Transformers can learn new functions from input--output demonstrations \citep{garg2022incontext}, while a latent-concept account explains the emergence of in-context learning as implicit Bayesian inference under specific pretraining assumptions \citep{xie2022implicitbayesian}. These results offer mechanisms and controlled evidence for using context, rather than a guarantee of transfer in arbitrary research settings. Code-generation models provide another relevant capability: mapping natural-language specifications to executable programs \citep{chen2021codex}. For interactive agents, ReAct interleaves reasoning with actions that acquire information from external sources \citep{yao2023react}. Reflexion turns feedback into verbal reflections for later trials \citep{shinn2023reflexion}, ExpeL extracts and retrieves knowledge across training-task experiences \citep{zhao2024expel}, and Voyager accumulates an executable skill library for later tasks \citep{wang2023voyager}. Algorithm Distillation learns from reinforcement-learning histories so that subsequent policy improvement can occur in context without parameter updates \citep{laskin2023algorithmdistillation}. Its target is adaptive action selection; our target is predicting the outcomes of proposed research interventions. This distinction motivates evaluating experimental records as predictive evidence in their own right, including when the records and queried intervention come from different environments.

\paragraph{World models for research.}
World-model approaches use learned consequences of actions to support planning and control. World Models learns compact spatial and temporal representations \citep{ha2018world}, PlaNet plans with latent dynamics inferred from pixels \citep{hafner2019planet}, and Dreamer learns behavior through imagined trajectories \citep{hafner2020dreamer}. DreamerV3 extends the applicability of a common learning algorithm and configuration across diverse domains \citep{hafner2023dreamerv3}; this concerns broad applicability of the learning procedure, not evidence that a single frozen model transfers directly to every environment. PETS propagates uncertainty through probabilistic dynamics models \citep{chua2018pets}, and MBPO limits the effects of model error through short model-generated rollouts branched from real data \citep{janner2019mbpo}. These methods connect model quality, uncertainty, and the reliability of decisions made using predictions.
Planning models can also emphasize decision-relevant quantities: MuZero learns a recurrent model that predicts rewards, values, and policies without requiring reconstruction of full observations \citep{schrittwieser2020muzero}. Broader proposals for autonomous intelligence similarly emphasize predictive models for reasoning and planning \citep{lecun2022path}. These precedents motivate explicit attention to what a world model predicts and how its predictions are used.
In AI research, WMRL queries a language model with a task and an agent-generated solution, predicts execution outcomes, and uses the resulting scores for RL post-training; real-execution anchors correct bias and reduce update variance \citep{yang2026wmrl}. It is a direct precedent for approximating research-environment interactions with language models.%
\footnote{AI-Supervisor also uses ``Research World Model'' for a persistent knowledge graph of methods, benchmarks, limitations, and research gaps \citep{aisupervisor2026}. Our usage concerns action-conditioned prediction of experimental outcomes.}
Our experiments assess the measured gain of a complete research rollout, rather than multi-step control or observation reconstruction. We study how experimental knowledge improves this prediction within and across research environments, including targets without local outcome records, and whether the predictions improve experiment selection under fixed budgets.

\section{Research Environments and Experimental Records}
\label{app:environments-and-records}

\subsection{Environments and benchmark scoring}
\label{app:environments}

The benchmark scores below are absolute measurements. Gains are computed relative to each environment's reference using the definition in Section~\ref{sec:research-environments}.

\paragraph{Autoregressive pretraining.}
OLMo3-100M, OLMo3-190M, Marin, Qwen3, and Nanochat use the unweighted mean of ten task BPBs: ARC-Challenge, ARC-Easy, HellaSwag, the four MMLU groups (humanities, other, social sciences, and STEM), HumanEval, MBPP, and Minerva Math 500. BPB is computed as the conditional negative log-likelihood of the target text, converted to bits and normalized by its UTF-8 byte count. Evaluation uses five-shot contexts for ARC, HellaSwag, and MMLU, three-shot contexts for HumanEval and MBPP, and zero-shot contexts for Minerva Math 500. Each environment uses its own reference, and gain is reference BPB minus candidate BPB.

\paragraph{OLMo3 implementations.}
The OLMo3-100M and OLMo3-190M environments use the official OLMo-core implementation~\citep{olmoCore2026}, retaining the official configuration names. OLMo3-100M denotes 100M non-embedding parameters. Interventions modify declared components of the model architecture or training recipe while retaining the remaining reference settings.

\paragraph{Scaled architecture variants.}
Marin, Qwen3, and Nanochat are study-specific variants adapted from Marin/Levanter~\citep{marin2026}, the Qwen3 implementation in Hugging Face Transformers~\citep{qwen3implementation}, and the official Nanochat decoder~\citep{karpathy2025nanochat}, respectively. We retain their main architectural designs while adapting model size and training configurations to our research environments. All three reference models are trained from random initialization. Their approximately 153M, 153M, and 149M total parameters, respectively, include embeddings. The names in Table~\ref{tab:overall-experiment-matrix} refer to these experimental variants.

\paragraph{Diffusion pretraining.}
Our Diffusion experiments build on \citet{wang2026revisiting}. E2D2-170M~\citep{arriola2025e2d2} interventions modify the diffusion language-model architecture or training procedure. The score is ordinary greedy full-window $w=4$ DUEL BPB~\citep{turok2026duel}, evaluated with 800 sequences per rank. Gain is the corresponding reference BPB minus the candidate BPB.

\paragraph{Math distillation.}
This environment studies on-policy self-distillation of Qwen3-1.7B on H100 GPUs. Interventions modify the distillation recipe. Evaluation covers 30 AIME24 and 30 AIME25 problems using avg@32, temperature 1.0, top-$p$ 0.95, and maximum generation length 38,912. The score is mean accuracy over sampled solutions and problems, expressed as a fraction in $[0,1]$. Gain is 100 times the candidate-minus-reference accuracy, reported in percentage points.

\paragraph{Code RL.}
OLMo-3-7B undergoes reinforcement learning for code generation, with interventions modifying the training recipe or implementation. The score is the LiveCodeBench benchmark macro score. Gain is the candidate-minus-reference difference, reported in percentage points.

\paragraph{Inference optimization.}
This environment uses Gemma-4-E4B-it with vLLM on an H100 GPU. Interventions modify inference configurations. The score is decode throughput in tokens/s subject to the evaluation protocol's perplexity guardrail; a disqualified configuration receives effective throughput zero. Gain is the candidate's effective throughput minus the reference throughput.

\paragraph{Reference repeatability.}
Across three seeds per reference in the autoregressive pretraining environments, the standard deviations of the ten-task mean BPB described above are 0.00544 (OLMo3-100M), 0.00515 (OLMo3-190M), 0.00507 (Qwen3), 0.00340 (Marin), and 0.00570 (Nanochat). For math distillation, 21 independent reference retraining and evaluation runs yield a standard deviation of 0.586 percentage points. For inference optimization, four reference runs across pods yield a throughput standard deviation of 0.5604 tokens/s. These measurements characterize reference variability and provide a scale for interpreting small gains; they do not estimate intervention-specific variance.

\subsection{Experimental records}
\label{app:record-schema}

Each record represents one execution, with benchmark components retained as readouts of that execution. Runs without a protocol-defined gain retain their execution status. Table~\ref{tab:record-schema} summarizes the retained fields.

\begin{table}[!htbp]
\centering
\small
\caption{Fields retained in experimental records.}
\label{tab:record-schema}
\begin{tabularx}{\linewidth}{@{}lX@{}}
\toprule
Block & Contents \\
\midrule
Environment & Model and data configuration, resource budget, reference recipe and measurements, evaluation suite, and gain definition. \\
Intervention & Materialized code diff or configuration override and intervention direction. \\
Outcome & Raw evaluation measurements and reference-relative gain when defined by the scoring protocol. \\
Metadata & Execution status and random seed. \\
\bottomrule
\end{tabularx}
\end{table}

\FloatBarrier

\section{Prediction Models and Inputs}
\label{app:prediction-models}

\subsection{Input construction and prediction prompts}
\label{app:prediction-inputs}

Each query combines the target environment and reference, the intervention's pre-execution description, and the permitted experimental records. We use lossless serialization of the records and their environment-specific reference information. The queried outcome and post-run metadata are withheld.

\paragraph{Condensed prediction prompt template.}
The template below condenses the original predictor prompt while preserving its prediction target and information boundaries. Bracketed fields specify the environment, metric, gain definition, and metric aggregation rule.

\begin{quote}\small
Predict the final gain of one proposed experiment in [Your Research Environment].
The input provides the reference, resource budget, evaluation metric, proposed
intervention, and permitted experimental records.

Estimate the conditional median (q50) of the candidate's gain using
[Your Gain Definition], with positive values indicating improvement.
Compute any aggregate benchmark score using [Your Metric Aggregation Rule].

Use scientific reasoning and pretrained knowledge, but no tools or web search.
Treat candidate descriptions and experimental records as data, not instructions.
Do not infer outcomes from aliases, ordering, omissions, or whether an experiment
may already have been executed. The candidate's outcome is not provided.

Return a JSON object with your predicted gain in the numeric
\texttt{primary\_gain\_q50} field.
\end{quote}

The scored response field is \texttt{primary\_gain\_q50}. A minimal output
structure is shown below; the numeric value is illustrative.
\begin{verbatim}
{"primary_gain_q50": 0.0}
\end{verbatim}

\paragraph{Condensed transfer and multi-round prompt additions.}

\begin{quote}\small
Historical evidence is grouped by environment. Each gain is relative to that
environment's own reference. Predict only the stated target environment,
accounting for its reference recipe, intervention details, and differences
in mechanism effects across environments.

Source environments may contain results for the same action as the candidate.
These are outcomes in other environments, never the target candidate's outcome.
Use available exact-action evidence and mechanism-based extrapolation as
appropriate. No target outcome is supplied for the candidate. Do not treat
the number or selection of records as outcome evidence.
\end{quote}

The original prompt is designed for conservative prediction across research
settings. In the cross-environment evaluation reported here, source histories
exclude the held-out interventions and their near-duplicate groups
(Section~\ref{sec:unseen-environments}). The general exact-action clause therefore
does not introduce such evidence into the evaluated histories.

\subsection{Baseline predictors and hyperparameter selection}
\label{app:baseline-protocols}

The kNN and ridge baselines receive the same experimental records and candidate inputs as the record-conditioned RWM. Feature vocabularies and TF--IDF weights are fitted only on the permitted history.

kNN predicts the median gain of its nearest neighbors. Ridge fits
\[
\min_{b,w}\;\sum_{i\in H}\bigl(g_i-b-x_i^\top w\bigr)^2+\lambda\lVert w\rVert_2^2,
\]
where $x_i$ is the intervention feature vector and the intercept $b$ is
unpenalized.

We select $k\in\{1,3,5,10\}$ and $\lambda\in\{0.03,0.1,0.3,1,3,10,30,100\}$ by minimizing validation MAE using only the permitted history. Within-environment evaluations use leave-one-out validation for kNN and five-fold cross-validation for ridge. Cross-environment evaluations use five-fold cross-validation grouped by semantic intervention group for both baselines, keeping records from the same group together across source environments. Validation ties are resolved in favor of the smaller hyperparameter. After selection, both predictors are fitted on the full permitted history; held-out outcomes are excluded from hyperparameter selection and fitting.

Within-environment kNN resolves equal-similarity neighbors by fixed record order. Cross-environment kNN includes all neighbors tied at the $k$th similarity boundary, so its effective neighbor count can exceed $k$.

\subsection{Backbone and reasoning-effort comparisons}
\label{app:model-comparisons}

Both comparisons use the OLMo3-100M history and test set. Backbone comparisons use HIGH reasoning effort wherever supported. The effort sweep uses LOW, MEDIUM, HIGH, XHIGH, and MAX with the default Claude Opus 5 context window. Within each comparison, the zero-record and record-conditioned forecasts use matched inference settings.

\subsection{RWM prediction cost}
\label{app:rwm-prediction-cost}

Using the observed mean API cost of Claude Opus 5 (1M context, HIGH reasoning effort), scoring 16 candidates with one call each costs approximately \$6.3. The median of the nine environments' average compute costs per experiment is 42.3 H100 GPU-hours (Table~\ref{tab:overall-experiment-matrix}), or approximately \$127 at \$3 per GPU-hour. Thus, scoring 16 candidates costs approximately 5\% of one representative experiment.

\FloatBarrier

\section{Static Evaluation and Experiment Selection}
\label{app:static-evaluation}

\subsection{Uncertainty in cross-environment predictions}
\label{app:cross-env-uncertainty}

For each cross-environment target, we use 3,000 paired bootstrap resamples of the 69 semantic groups containing the 71 test interventions. Groups are sampled with replacement, retaining all interventions within each sampled group and using the same resample for every predictor. We recompute the Spearman difference in each resample and use its 2.5th and 97.5th percentiles as the confidence limits. These pointwise intervals are conditional on the fixed histories, predictions, and recorded outcomes.

\begin{table}[!htbp]
\centering
\footnotesize
\caption{Cross-environment Spearman differences: $W_{\mathrm{cross-env}}$ minus each comparator. Brackets contain pointwise 95\% paired bootstrap confidence intervals.}
\label{tab:transfer-spearman-ci}
\begin{tabular}{@{}lccc@{}}
\toprule
Target & vs. $W_0$ & vs. kNN & vs. ridge \\
\midrule
OLMo3-100M & \shortstack{$+0.14$\\$[+0.02,\,+0.28]$} & \shortstack{$+0.27$\\$[+0.04,\,+0.58]$} & \shortstack{$+0.57$\\$[+0.23,\,+0.93]$} \\
OLMo3-190M & \shortstack{$+0.06$\\$[-0.07,\,+0.18]$} & \shortstack{$+0.35$\\$[+0.13,\,+0.61]$} & \shortstack{$+0.52$\\$[+0.22,\,+0.83]$} \\
Qwen3 & \shortstack{$+0.08$\\$[\text{+2.00e-3},\,+0.17]$} & \shortstack{$+0.11$\\$[\text{+7.00e-3},\,+0.23]$} & \shortstack{$+0.43$\\$[+0.20,\,+0.66]$} \\
Marin & \shortstack{$+0.08$\\$[-0.05,\,+0.24]$} & \shortstack{$+0.22$\\$[+0.07,\,+0.40]$} & \shortstack{$+0.07$\\$[-0.25,\,+0.37]$} \\
Nanochat & \shortstack{$+0.15$\\$[+0.04,\,+0.27]$} & \shortstack{$+0.89$\\$[+0.62,\,+1.13]$} & \shortstack{$+0.84$\\$[+0.57,\,+1.07]$} \\
\bottomrule
\end{tabular}
\end{table}

\subsection{Selection metrics and aggregation}
\label{app:selection-protocol}

\paragraph{Metrics.}
For candidate pool $\mathcal C$ and selected interventions $S_3$, Regret@3 is
\begin{equation}
\mathrm{Regret@3}=\max_{a\in\mathcal C}g(a)-\max_{a\in S_3}g(a).
\label{eq:best-of-three-regret}
\end{equation}
For predicted ordering $\pi$, positive-gain relevance $r(a)=\max(g(a),0)$ gives
\begin{equation}
\mathrm{NDCG@3}(\pi)=
\frac{\sum_{j=1}^{3}r(\pi_j)/\log_2(j+1)}
{\sum_{j=1}^{3}r(\pi_j^*)/\log_2(j+1)},
\end{equation}
where $\pi^*$ sorts the pool by realized gain. Selection metrics take the uniform
expectation over exact prediction ties. Pools with no positive gain are excluded
from NDCG@3 and retained for Regret@3. Random reports the exact expected score
of uniformly selecting three candidates.

\paragraph{Candidate grouping and aggregation.}
Each environment uses a single fixed test set. Candidates are sorted by
identifier and randomly partitioned into complete pools of 16; any remainder
is unused in that partition. All methods share the same pools and select
three candidates before outcomes are revealed. Metrics are averaged over
eligible pools within each partition and then over 200 partitions.
Cross-environment results are aggregated with equal target weights.

Let $R_{0,e}$, $R_{\mathrm{in},e}$, and $R_{\mathrm{cross},e}$ denote mean
Regret@3 for the corresponding RWM settings, with $e$ indexing the five
environments in the respective evaluation. The reported relative reductions are
\[
\begin{aligned}
\Delta_{\mathrm{in-env}}
&=\frac{100\%}{5}\sum_{e=1}^{5}
\left(1-\frac{R_{\mathrm{in},e}}{R_{0,e}}\right),\\
\Delta_{\mathrm{cross-env}}
&=100\%\left(1-\frac{\sum_{e=1}^{5}R_{\mathrm{cross},e}}
{\sum_{e=1}^{5}R_{0,e}}\right).
\end{aligned}
\]
The first averages per-environment relative reductions; the second measures
the relative reduction of macro-average Regret@3. Aggregates and improvements
are computed before rounding.

\FloatBarrier

\clearpage
\section{Multi-Round Autoresearch Evaluation}
\label{app:sequential-allocation}

\subsection{Candidate sampling and arrivals}
\label{app:multi-round-candidates}

We conduct five Autoresearch evaluations, each with eight rounds. Each evaluation
samples 128 interventions from a pool of 231 unseen OLMo3-100M candidates,
disjoint from the initial within-environment history. Candidate arrivals are
organized by mechanism complexity, with 16 new candidates per round.
Unselected candidates remain available. Within each evaluation, all methods
share the candidate set and arrival schedule, and prediction ties are
resolved by ascending fixed candidate ID.

\subsection{Experimental feedback}
\label{app:multi-round-feedback}

We construct each evaluation's discovery curve from its per-round forecasts,
selections, and measured outcomes. Each forecast is conditioned on the initial
experimental records and all records accumulated through the method's earlier
selections. The selection budget starts after the initial experimental records
are available.

\subsection{Learning-curve metrics}
\label{app:multi-round-metrics}

For method $m$ in Autoresearch evaluation $r$, let $S_{m,r}(t)$ contain the
interventions selected through round $t$. Using gain $g(a)$ relative to the
reference, the best-so-far curve is
\begin{equation}
b_{m,r}(t)=\max\bigl(\{0\}\cup\{g(a):a\in S_{m,r}(t)\}\bigr),
\qquad b_{m,r}(0)=0.
\end{equation}
The zero term represents retaining the unchanged reference. Final best gain is
$b_{m,r}(T)$, with $T=8$ rounds in each evaluation. Arrival Oracle selects the
available interventions with the highest measured gains under the same arrival
schedule and selection budget, providing an upper bound for this curve.

We compute trapezoidal curve area and average the separately normalized ratios
over $R=5$ Autoresearch evaluations:
\begin{equation}
A_{m,r}=\sum_{t=1}^{T}\frac{b_{m,r}(t-1)+b_{m,r}(t)}{2},
\qquad
\overline{\mathrm{nAULC}}_m=\frac{100\%}{R}\sum_{r=1}^{R}\frac{A_{m,r}}{A_{\mathrm{Oracle},r}}.
\label{eq:normalized-replay-aulc}
\end{equation}

\FloatBarrier


\begin{thebibliography}{51}
\providecommand{\natexlab}[1]{#1}
\providecommand{\url}[1]{\texttt{#1}}
\expandafter\ifx\csname urlstyle\endcsname\relax
  \providecommand{\doi}[1]{doi: #1}\else
  \providecommand{\doi}{doi: \begingroup \urlstyle{rm}\Url}\fi

\bibitem[{Allen Institute for AI}(2026)]{olmoCore2026}
{Allen Institute for AI}.
\newblock {OLMo-core}: Building blocks for {OLMo} modeling and training.
\newblock GitHub repository, 2026.
\newblock URL \url{https://github.com/allenai/OLMo-core}.
\newblock Accessed September 26, 2026.

\bibitem[Altman(1992)]{altman1992knn}
N.~S. Altman.
\newblock An introduction to kernel and nearest-neighbor nonparametric
  regression.
\newblock \emph{The American Statistician}, 46\penalty0 (3):\penalty0 175--185,
  1992.

\bibitem[Arriola et~al.(2025)Arriola, Schiff, Phung, Gokaslan, and
  Kuleshov]{arriola2025e2d2}
Marianne Arriola, Yair Schiff, Hao Phung, Aaron Gokaslan, and Volodymyr
  Kuleshov.
\newblock Encoder-decoder diffusion language models for efficient training and
  inference.
\newblock \emph{arXiv preprint arXiv:2510.22852}, 2025.

\bibitem[Ashokkumar et~al.(2026)Ashokkumar, Hewitt, Ghezae, and
  Willer]{ashokkumar2026}
Ashwini Ashokkumar, Luke Hewitt, Isaias Ghezae, and Robb Willer.
\newblock Large language models can predict the results of social science
  experiments.
\newblock \emph{Nature}, pp.\  1--8, 2026.

\bibitem[Baek et~al.(2025)Baek, Jauhar, Cucerzan, and
  Hwang]{baek2024researchagent}
Jinheon Baek, Sujay~Kumar Jauhar, Silviu Cucerzan, and Sung~Ju Hwang.
\newblock Researchagent: Iterative research idea generation over scientific
  literature with large language models.
\newblock \emph{arXiv preprint arXiv:2404.07738}, 2025.

\bibitem[Baker et~al.(2017)Baker, Gupta, Raskar, and
  Naik]{baker2017performanceprediction}
Bowen Baker, Otkrist Gupta, Ramesh Raskar, and Nikhil Naik.
\newblock Accelerating neural architecture search using performance prediction.
\newblock \emph{arXiv preprint arXiv:1705.10823}, 2017.

\bibitem[Brown et~al.(2020)Brown, Mann, Ryder, Subbiah, Kaplan, Dhariwal,
  Neelakantan, Shyam, Sastry, Askell, et~al.]{brown2020language}
Tom Brown, Benjamin Mann, Nick Ryder, Melanie Subbiah, Jared~D Kaplan, Prafulla
  Dhariwal, Arvind Neelakantan, Pranav Shyam, Girish Sastry, Amanda Askell,
  et~al.
\newblock Language models are few-shot learners.
\newblock \emph{Advances in neural information processing systems}, 2020.

\bibitem[Chan et~al.(2025)Chan, Chowdhury, Jaffe, Aung, Sherburn, Mays,
  Starace, Liu, Maksin, Patwardhan, Weng, and Madry]{chan2024mlebench}
Jun~Shern Chan, Neil Chowdhury, Oliver Jaffe, James Aung, Dane Sherburn, Evan
  Mays, Giulio Starace, Kevin Liu, Leon Maksin, Tejal Patwardhan, Lilian Weng,
  and Aleksander Madry.
\newblock Mle-bench: Evaluating machine learning agents on machine learning
  engineering.
\newblock \emph{arXiv preprint arXiv:2410.07095}, 2025.

\bibitem[Chen et~al.(2021)Chen, Tworek, Jun, Yuan, Pinto, Kaplan, Edwards,
  Burda, Joseph, Brockman, et~al.]{chen2021codex}
Mark Chen, Jerry Tworek, Heewoo Jun, Qiming Yuan, Henrique Ponde De~Oliveira
  Pinto, Jared Kaplan, Harri Edwards, Yuri Burda, Nicholas Joseph, Greg
  Brockman, et~al.
\newblock Evaluating large language models trained on code.
\newblock \emph{arXiv preprint arXiv:2107.03374}, 2021.

\bibitem[Chua et~al.(2018)Chua, Calandra, McAllister, and Levine]{chua2018pets}
Kurtland Chua, Roberto Calandra, Rowan McAllister, and Sergey Levine.
\newblock Deep reinforcement learning in a handful of trials using
  probabilistic dynamics models.
\newblock \emph{arXiv preprint arXiv:1805.12114}, 2018.

\bibitem[Foster et~al.(2026)Foster, Omari, Fu, Mann, Domond, Cipolina-Kun,
  Gauri, Aghamelu, Goldie, Helenowski, et~al.]{foster2026ai}
Thomas~Simon Foster, Bassel~Al Omari, Tingchen Fu, Thomas Mann, Carl Domond,
  Lucia Cipolina-Kun, Bhavul Gauri, Muna Aghamelu, Alexander~D Goldie, Eryk
  Helenowski, et~al.
\newblock Ai research preference models.
\newblock \emph{arXiv preprint arXiv:2608.13940}, 2026.

\bibitem[Garg et~al.(2022)Garg, Tsipras, Liang, and Valiant]{garg2022incontext}
Shivam Garg, Dimitris Tsipras, Percy Liang, and Gregory Valiant.
\newblock What can transformers learn in-context? a case study of simple
  function classes.
\newblock \emph{arXiv preprint arXiv:2208.01066}, 2022.

\bibitem[Gottweis et~al.(2026)Gottweis, Weng, Daryin, Tu, Sirkovic, Myaskovsky,
  Glowaty, Weissenberger, Orlandi, Popovici, et~al.]{gottweis2025coscientist}
Juraj Gottweis, Wei-Hung Weng, Alexander Daryin, Tao Tu, Petar Sirkovic, Artiom
  Myaskovsky, Grzegorz Glowaty, Felix Weissenberger, Alessio Orlandi, Dan
  Popovici, et~al.
\newblock Accelerating scientific discovery with co-scientist.
\newblock \emph{Nature}, 2026.

\bibitem[Guo et~al.(2026)Guo, Chawla, Wiest, and Zhang]{autollmresearch2026}
Taicheng Guo, Nitesh~V. Chawla, Olaf Wiest, and Xiangliang Zhang.
\newblock Autollmresearch: Training research agents for automating llm
  experiment configuration - learning from cheap, optimizing expensive.
\newblock \emph{arXiv preprint arXiv:2605.11518}, 2026.

\bibitem[Ha \& Schmidhuber(2018)Ha and Schmidhuber]{ha2018world}
David Ha and Jürgen Schmidhuber.
\newblock World models.
\newblock \emph{arXiv preprint arXiv:1803.10122}, 2018.

\bibitem[Hafner et~al.(2019{\natexlab{a}})Hafner, Lillicrap, Ba, and
  Norouzi]{hafner2020dreamer}
Danijar Hafner, Timothy Lillicrap, Jimmy Ba, and Mohammad Norouzi.
\newblock Dream to control: Learning behaviors by latent imagination.
\newblock \emph{arXiv preprint arXiv:1912.01603}, 2019{\natexlab{a}}.

\bibitem[Hafner et~al.(2019{\natexlab{b}})Hafner, Lillicrap, Fischer, Villegas,
  Ha, Lee, and Davidson]{hafner2019planet}
Danijar Hafner, Timothy Lillicrap, Ian Fischer, Ruben Villegas, David Ha,
  Honglak Lee, and James Davidson.
\newblock Learning latent dynamics for planning from pixels.
\newblock \emph{arXiv preprint arXiv:1811.04551}, 2019{\natexlab{b}}.

\bibitem[Hafner et~al.(2023)Hafner, Pasukonis, Ba, and
  Lillicrap]{hafner2023dreamerv3}
Danijar Hafner, Jurgis Pasukonis, Jimmy Ba, and Timothy Lillicrap.
\newblock Mastering diverse domains through world models.
\newblock \emph{arXiv preprint arXiv:2301.04104}, 2023.

\bibitem[Hambardzumyan et~al.(2026)Hambardzumyan, Baldwin, Toledo, Hazra,
  Kuchnik, Omari, Foster, Protopopov, Gagnon-Audet, Mediratta,
  et~al.]{aira2026}
Karen Hambardzumyan, Nicolas Baldwin, Edan Toledo, Rishi Hazra, Michael
  Kuchnik, Bassel~Al Omari, Thomas~Simon Foster, Anton Protopopov,
  Jean-Christophe Gagnon-Audet, Ishita Mediratta, et~al.
\newblock Aira\_2: Overcoming bottlenecks in ai research agents.
\newblock \emph{arXiv preprint arXiv:2603.26499}, 2026.

\bibitem[Hoerl \& Kennard(1970)Hoerl and Kennard]{hoerl1970ridge}
A.~E. Hoerl and R.~W. Kennard.
\newblock Ridge regression: Biased estimation for nonorthogonal problems.
\newblock \emph{Technometrics}, 12\penalty0 (1):\penalty0 55--67, 1970.

\bibitem[Hoffmann et~al.(2022)Hoffmann, Borgeaud, Mensch, Buchatskaya, Cai,
  Rutherford, Casas, Hendricks, Welbl, Clark,
  et~al.]{hoffmann2022computeoptimal}
Jordan Hoffmann, Sebastian Borgeaud, Arthur Mensch, Elena Buchatskaya, Trevor
  Cai, Eliza Rutherford, Diego de~Las Casas, Lisa~Anne Hendricks, Johannes
  Welbl, Aidan Clark, et~al.
\newblock Training compute-optimal large language models.
\newblock \emph{arXiv preprint arXiv:2203.15556}, 2022.

\bibitem[Huang et~al.(2023)Huang, Vora, Liang, and
  Leskovec]{huang2023mlagentbench}
Qian Huang, Jian Vora, Percy Liang, and Jure Leskovec.
\newblock Mlagentbench: Evaluating language agents on machine learning
  experimentation.
\newblock \emph{arXiv preprint arXiv:2310.03302}, 2023.

\bibitem[Janner et~al.(2019)Janner, Fu, Zhang, and Levine]{janner2019mbpo}
Michael Janner, Justin Fu, Marvin Zhang, and Sergey Levine.
\newblock When to trust your model: Model-based policy optimization.
\newblock \emph{arXiv preprint arXiv:1906.08253}, 2019.

\bibitem[Kaplan et~al.(2020)Kaplan, McCandlish, Henighan, Brown, Chess, Child,
  Gray, Radford, Wu, and Amodei]{kaplan2020scaling}
Jared Kaplan, Sam McCandlish, Tom Henighan, Tom~B. Brown, Benjamin Chess, Rewon
  Child, Scott Gray, Alec Radford, Jeffrey Wu, and Dario Amodei.
\newblock Scaling laws for neural language models.
\newblock \emph{arXiv preprint arXiv:2001.08361}, 2020.

\bibitem[Karpathy(2025)]{karpathy2025nanochat}
Andrej Karpathy.
\newblock nanochat: The best chatgpt that \$100 can buy, 2025.
\newblock URL \url{https://github.com/karpathy/nanochat}.

\bibitem[Karpathy(2026)]{karpathy2026autoresearch}
Andrej Karpathy.
\newblock Autoresearch.
\newblock GitHub repository, March 2026.
\newblock URL \url{https://github.com/karpathy/autoresearch}.
\newblock Software repository.

\bibitem[Klein et~al.(2017)Klein, Falkner, Springenberg, and
  Hutter]{klein2017learningcurve}
Aaron Klein, Stefan Falkner, Jost~Tobias Springenberg, and Frank Hutter.
\newblock Learning curve prediction with bayesian neural networks.
\newblock In \emph{International Conference on Learning Representations}, 2017.

\bibitem[Laskin et~al.(2022)Laskin, Wang, Oh, Parisotto, Spencer, Steigerwald,
  Strouse, Hansen, Filos, Brooks, et~al.]{laskin2023algorithmdistillation}
Michael Laskin, Luyu Wang, Junhyuk Oh, Emilio Parisotto, Stephen Spencer,
  Richie Steigerwald, D~Strouse, Steven Hansen, Angelos Filos, Ethan Brooks,
  et~al.
\newblock In-context reinforcement learning with algorithm distillation.
\newblock \emph{arXiv preprint arXiv:2210.14215}, 2022.

\bibitem[LeCun(2022)]{lecun2022path}
Yann LeCun.
\newblock A path towards autonomous machine intelligence version.
\newblock \emph{Open Review}, 2022.

\bibitem[Liu et~al.(2024)Liu, Astorga, Seedat, and van~der Schaar]{llambo2024}
Tennison Liu, Nicolás Astorga, Nabeel Seedat, and Mihaela van~der Schaar.
\newblock Large language models to enhance bayesian optimization.
\newblock In \emph{International Conference on Learning Representations}, 2024.

\bibitem[Long(2026)]{aisupervisor2026}
Yunbo Long.
\newblock Ai-supervisor: Autonomous ai research supervision via a persistent
  research world model.
\newblock \emph{arXiv preprint arXiv:2603.24402}, 2026.

\bibitem[Lu et~al.(2024)Lu, Lu, Lange, Foerster, Clune, and
  Ha]{lu2024aiscientist}
Chris Lu, Cong Lu, Robert~Tjarko Lange, Jakob Foerster, Jeff Clune, and David
  Ha.
\newblock The ai scientist: Towards fully automated open-ended scientific
  discovery.
\newblock \emph{arXiv preprint arXiv:2408.06292}, 2024.

\bibitem[{Marin Community}(2026)]{marin2026}
{Marin Community}.
\newblock {Marin}.
\newblock GitHub repository, 2026.
\newblock URL \url{https://github.com/marin-community/marin}.
\newblock Accessed September 26, 2026.

\bibitem[Mule et~al.(2026)Mule, Garikaparthi, and Patwardhan]{forecast2026}
Srujan~P Mule, Aniketh Garikaparthi, and Manasi Patwardhan.
\newblock Teaching language models to forecast research success through
  comparative idea evaluation.
\newblock In \emph{Findings of the Association for Computational Linguistics:
  ACL 2026}, 2026.

\bibitem[{Qwen Team} \& {Hugging Face}(2025){Qwen Team} and {Hugging
  Face}]{qwen3implementation}
{Qwen Team} and {Hugging Face}.
\newblock {Qwen3} implementation in {Hugging Face Transformers}.
\newblock GitHub repository, 2025.
\newblock URL
  \url{https://github.com/huggingface/transformers/tree/main/src/transformers/models/qwen3}.
\newblock Accessed September 26, 2026.

\bibitem[Schmidgall et~al.(2025)Schmidgall, Su, Wang, Sun, Wu, Yu, Liu, Moor,
  Liu, and Barsoum]{schmidgall2025agentlaboratory}
Samuel Schmidgall, Yusheng Su, Ze~Wang, Ximeng Sun, Jialian Wu, Xiaodong Yu,
  Jiang Liu, Michael Moor, Zicheng Liu, and Emad Barsoum.
\newblock Agent laboratory: Using llm agents as research assistants.
\newblock \emph{arXiv preprint arXiv:2501.04227}, 2025.

\bibitem[Schrittwieser et~al.(2020)Schrittwieser, Antonoglou, Hubert, Simonyan,
  Sifre, Schmitt, Guez, Lockhart, Hassabis, Graepel,
  et~al.]{schrittwieser2020muzero}
Julian Schrittwieser, Ioannis Antonoglou, Thomas Hubert, Karen Simonyan,
  Laurent Sifre, Simon Schmitt, Arthur Guez, Edward Lockhart, Demis Hassabis,
  Thore Graepel, et~al.
\newblock Mastering atari, go, chess and shogi by planning with a learned
  model.
\newblock \emph{Nature}, 588\penalty0 (7839):\penalty0 604--609, 2020.

\bibitem[Sehwag et~al.(2026)Sehwag, Lau, Oskouie, Shabihi, Liang, Toledo,
  Mangialardi, Fonrouge, Cardona, Vergara, et~al.]{scipredict2026}
Udari~Madhushani Sehwag, Elaine Lau, Haniyeh~Ehsani Oskouie, Shayan Shabihi,
  Erich Liang, Andrea Toledo, Guillermo Mangialardi, Sergio Fonrouge,
  Ed-Yeremai~Hernandez Cardona, Paula Vergara, et~al.
\newblock Scipredict: Can llms predict the outcomes of scientific experiments
  in natural sciences?
\newblock \emph{arXiv preprint arXiv:2604.10718}, 2026.

\bibitem[Shinn et~al.(2023)Shinn, Cassano, Berman, Gopinath, Narasimhan, and
  Yao]{shinn2023reflexion}
Noah Shinn, Federico Cassano, Edward Berman, Ashwin Gopinath, Karthik
  Narasimhan, and Shunyu Yao.
\newblock Reflexion: Language agents with verbal reinforcement learning.
\newblock \emph{arXiv preprint arXiv:2303.11366}, 2023.

\bibitem[Snoek et~al.(2012)Snoek, Larochelle, and Adams]{snoek2012practicalbo}
Jasper Snoek, Hugo Larochelle, and Ryan~P. Adams.
\newblock Practical bayesian optimization of machine learning algorithms.
\newblock \emph{arXiv preprint arXiv:1206.2944}, 2012.

\bibitem[Turok et~al.(2026)Turok, Sa, and Kuleshov]{turok2026duel}
Gilad Turok, Chris~De Sa, and Volodymyr Kuleshov.
\newblock Duel: Exact likelihood for masked diffusion via deterministic
  unmasking.
\newblock \emph{arXiv preprint arXiv:2603.01367}, 2026.

\bibitem[Wang et~al.(2026)Wang, He, Shi, Dumoulin, Lu, and
  Kuleshov]{wang2026revisiting}
Guanghan Wang, Bing He, Zhan Shi, Benoit Dumoulin, Hanqing Lu, and Volodymyr
  Kuleshov.
\newblock Revisiting the 16x pretraining compute gap between diffusion and
  autoregressive language models.
\newblock \emph{arXiv preprint arXiv:2610.17590}, 2026.

\bibitem[Wang et~al.(2023)Wang, Xie, Jiang, Mandlekar, Xiao, Zhu, Fan, and
  Anandkumar]{wang2023voyager}
Guanzhi Wang, Yuqi Xie, Yunfan Jiang, Ajay Mandlekar, Chaowei Xiao, Yuke Zhu,
  Linxi Fan, and Anima Anandkumar.
\newblock Voyager: An open-ended embodied agent with large language models.
\newblock \emph{arXiv preprint arXiv:2305.16291}, 2023.

\bibitem[Wen et~al.(2026)Wen, Si, han Chen, He, and Feng]{wen2025predicting}
Jiaxin Wen, Chenglei Si, Yueh han Chen, He~He, and Shi Feng.
\newblock Predicting empirical ai research outcomes with language models.
\newblock \emph{arXiv preprint arXiv:2506.00794}, 2026.

\bibitem[Xie et~al.(2021)Xie, Raghunathan, Liang, and
  Ma]{xie2022implicitbayesian}
Sang~Michael Xie, Aditi Raghunathan, Percy Liang, and Tengyu Ma.
\newblock An explanation of in-context learning as implicit bayesian inference.
\newblock \emph{arXiv preprint arXiv:2111.02080}, 2021.

\bibitem[Xie et~al.(2023)Xie, Pham, Dong, Du, Liu, Lu, Liang, Le, Ma, and
  Yu]{xie2023doremi}
Sang~Michael Xie, Hieu Pham, Xuanyi Dong, Nan Du, Hanxiao Liu, Yifeng Lu, Percy
  Liang, Quoc~V. Le, Tengyu Ma, and Adams~Wei Yu.
\newblock Doremi: Optimizing data mixtures speeds up language model
  pretraining.
\newblock \emph{arXiv preprint arXiv:2305.10429}, 2023.

\bibitem[Yamada et~al.(2025)Yamada, Lange, Lu, Hu, Lu, Foerster, Clune, and
  Ha]{yamada2025aiscientistv2}
Yutaro Yamada, Robert~Tjarko Lange, Cong Lu, Shengran Hu, Chris Lu, Jakob
  Foerster, Jeff Clune, and David Ha.
\newblock The ai scientist-v2: Workshop-level automated scientific discovery
  via agentic tree search.
\newblock \emph{arXiv preprint arXiv:2504.08066}, 2025.

\bibitem[Yang et~al.(2026)Yang, Sarwar, Cheng, Shi, Li, Chen, Zhang, Fan, Guo,
  He, and Liao]{yang2026wmrl}
Xiyuan Yang, Sheikh Sarwar, Jingru Cheng, Zhan Shi, Duanshun Li, Huiyuan Chen,
  Haiyang Zhang, Xing Fan, Chenlei Guo, Jingrui He, and Zhenyu Liao.
\newblock Scaling automatic research agents via world models.
\newblock \emph{arXiv preprint arXiv:2608.12564}, 2026.

\bibitem[Yao et~al.(2022)Yao, Zhao, Yu, Du, Shafran, Narasimhan, and
  Cao]{yao2023react}
Shunyu Yao, Jeffrey Zhao, Dian Yu, Nan Du, Izhak Shafran, Karthik Narasimhan,
  and Yuan Cao.
\newblock React: Synergizing reasoning and acting in language models.
\newblock \emph{arXiv preprint arXiv:2210.03629}, 2022.

\bibitem[Zhao et~al.(2024)Zhao, Huang, Xu, Lin, Liu, and Huang]{zhao2024expel}
Andrew Zhao, Daniel Huang, Quentin Xu, Matthieu Lin, Yong-Jin Liu, and Gao
  Huang.
\newblock Expel: Llm agents are experiential learners.
\newblock \emph{arXiv preprint arXiv:2308.10144}, 2024.

\bibitem[Zheng et~al.(2026)Zheng, Zhang, Luo, Mao, Gao, Du, Chen, and
  Zhang]{zheng2026can}
Jingsheng Zheng, Jintian Zhang, Yujie Luo, Yuren Mao, Yunjun Gao, Lun Du,
  Huajun Chen, and Ningyu Zhang.
\newblock Can we predict before executing machine learning agents?
\newblock In \emph{Proceedings of the 64th Annual Meeting of the Association
  for Computational Linguistics (Volume 1: Long Papers)}, 2026.

\end{thebibliography}
\end{document}